# Greedy Graph Searching for Vascular Tracking in Angiographic Image Sequences

Huihui Fang[1], Jian Yang[1,*], Jianjun Zhu[1], Danni Ai[1], Yong Huang[1], Yurong Jiang[1], Hong Song[2], Yongtian Wang[1]

[1]Beijing Engineering Research Center of Mixed Reality and Advanced Display, School of Optoelectronics, Beijing Institute of Technology, Beijing 100081, China

[2]School of Software, Beijing Institute of Technology, Beijing 100081, China

*Corresponding author:

Jian Yang, Email: jyang@bit.edu.cn

**Purpose:** Vascular tracking of angiographic image sequences is one of the most clinically important tasks in the diagnostic assessment and interventional guidance of cardiac disease due to providing dynamic structural information for precise motion analysis and 3D+t reconstruction. However, this task can be challenging to accomplish because of unsatisfactory angiography image quality and complex vascular structures. Thus, this study converted vascular tracking into branch matching and proposed a new greedy graph search-based method for it.

**Methods:** Each vascular branch was separated from the vasculature and was tracked independently. Then, all branches were combined using topology optimization, thereby resulting in complete vasculature tracking. An intensity-based image registration method was applied to determine the tracking range, and the deformation field between two consecutive frames was calculated. The vascular branch was described using a vascular centerline extraction method with multi-probability fusion-based topology optimization. We introduced an undirected acyclic graph establishment technique. A greedy search method was proposed to acquire all possible paths in the graph that might match the tracked vascular branch. The final tracking result was selected by branch matching using dynamic time warping with a DAISY descriptor.

**Results:** For single branch dataset SBD, the proposed method was evaluated on 12 angiographic image sequences with 77 angiograms of contrast agent-filled vessels. The average precision, sensitivity and F1 score of the tracking result of all angiograms were 0.90, 0.89 and 0.89, respectively. The average F1 score of the tracking results of the first, middle and last frames in all sequences were 0.91, 0.91 and 0.86, respectively. In the vessel tree dataset VTD, the proposed method was validated on 9 angiographic image sequences with 58 angiograms of contrast agent-filled vessels. The average precision, sensitivity and F1 score of the tracking result of all angiograms were 0.89, 0.87 and 0.88, respectively. The average F1 score of the tracking results of the first, middle and last frames in all sequences were 0.94, 0.88 and 0.82, respectively. Compared with five other state-of-the-art methods, the proposed method accurately tracked the vasculature from angiographic image sequences and the results were insignificantly affected by the tracking span in both datasets.

**Conclusions:** The solution to the problem reflected both the spatial and textural information between successive frames. Thus, the proposed method is robust and highly effective in vascular tracking of angiographic image sequences and the approach provided a universal solution to address the problem of filamentary structure tracking.

**Key words:** Angiographic image sequences, branch matching, dynamic time warping, graph searching, vasculature tracking.

# 1. Introduction

X-ray coronary angiography is an interventional imaging modality in which a contrast agent is injected into the blood vessel. Given that, X-ray cannot penetrate the contrast agent, the blood vessel state is observed based on the image displayed by the contrast agent under X-ray. Given its high resolution and real-time imaging, coronary angiography is considered the gold standard for diagnosing coronary arterial disease (CAD). Coronary angiography is the most commonly used and effective imaging modality in the treatment of CAD, including navigation for percutaneous coronary intervention [1, 2], providing vascular morphological information for diagnosing cardiovascular disease [3] and so on. The vascular structures in angiographic sequences are necessary to be extracted in interventional treatment guidance and post-processing analysis processes, such as quantitative measurement of 2D blood vessels [4], establishment of 3D vascular models [5-8], quantitative analysis of 3D blood vessels [9], and so on. What is more, coronary arteries cover the surface of the epicardium and periodically move with the heartbeat. Therefore, the analysis of vascular structures extracted from sequences can be used to perform cardiac cycle reconstruction [10] and cardiac dynamic analysis [11]. However, the extraction of vasculature from angiography sequences still present many challenges, such as low contrast in the angiogram, irregular vascular morphology, cross-vascular structure, and foreshortening.

Many existing approaches have focused on vasculature extraction from a single angiography image, including pixel-wise enhancement methods [12], energy optimization methods [13], learning-based methods [14], and our previous path searching-based method [15]. However, these methods do not consider temporal continuity and may therefore provide inconsistent vascular extraction results in a sequence. Fig. 1 shows the tracking of the vasculature in angiographic sequences, in which the target vascular structure in the first frame of a sequence is given either manually or automatically, and the vascular structures in the remaining frames are automatically tracked.

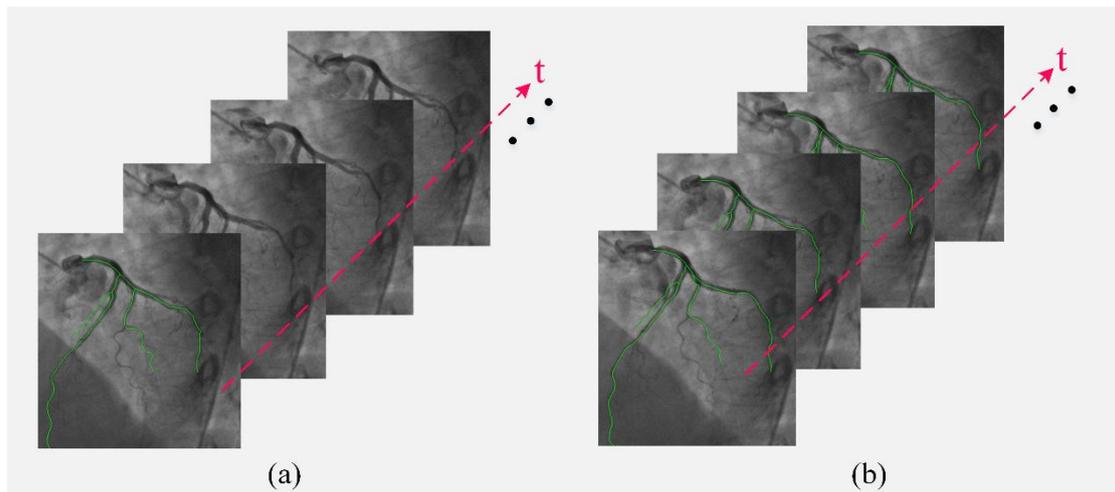

Fig.1 Overview of vascular tracking in angiographic image sequences. (a) An angiographic sequence and an initial vascular annotation (green) in the first frame. (b) Tracked vascular results in the remaining frames (green).

Existing tracking methods for the vasculature and interventional devices can be divided into three types: motion tracking-based methods, image registration-based methods, and prior model matching-based methods. In motion tracking-based methods, key points on the vasculature or interventional device are first extracted from the reference image. Then, the positions of the key points are tracked in the subsequent sequence images. Finally, a complete vasculature or interventional device is obtained by connecting the key points on all sequence images. Sun et al. [16] regarded the vascular centerline as a curve and improved the SNAKE model [17], which can be applied to the deformation of the open curve. Thus, they used this model to track the motion of points on the curve to implement vascular tracking. Cheng et al. [18] considered the vasculature as a collection of multi-target points and used low-rank tensor approximation with model propagation method to track the curvilinear structure. Chu et al. [19] first extracted vascular features by using deep learning method to obtain the possible positions of the vasculature in angiography sequences. Then, they tracked the vascular structures by utilizing a method similar to that described by Cheng et al. Zhang et al. [20] extracted representative target points on the vessels, such as bifurcation points, tracked the target points, and finally obtained the vascular structures on each image by connecting the target points. Shin et al. [21] used image registration method to determine the possible position of the vasculature in the succeeding frame and used a Markov model to optimize the position of the target point. Then, they applied FastMarching [22] and minimum path algorithm [23] to obtain the connection among target points and form a complete vascular structure. These motion tracking-based methods can easily fall into the local optimal value of the energy function because of the similarity of the vascular textures, resulting in the deviation of the final tracked vascular structures.

Image registration-based method utilizes the textural similarity of consecutive frames to calculate a deformation field and maps the vasculature or interventional devices to their corresponding positions. Compas et al. [24] used registration method for the first time to find the location of vascular structures on each frame and detect the coronary stenosis from angiography sequences. The vascular centerlines obtained by this method are inaccurate, unsmooth, off-center, and lacking the connection relationship among points. Existing prior model matching-based methods require a 3D model as reference, and the vascular structures tracked from angiography sequences are the projective positions of the deformation 3D model after 3D-2D vascular registration, in which the 2D centerlines are extracted from the angiographic sequences. One of the

3D-2D registration methods is the use of point cloud, in which the point set extracted from the 3D model is registered to the point set extracted from the 2D angiography image. Liu et al. [25] extracted candidate vascular structures from angiographic images and utilized iterative closest point technique to register the left anterior descending (LAD) in the 3D model and the candidate vascular structures in the 2D model to acquire the position of the LAD in 2D. Another 3D-2D registration method highlights the vascular topology. In this method, the vascular structures in two modalities are modeled as directed or undirected graphs. The points on vascular structures and the connections among points determine the nodes and edges in the graph, respectively. Holistic vascular registration is implemented by graph matching [26-28], which requires the topological structure of the 2D vasculature. However, the ideal vascular topology is difficult to obtain using automatic centerline extraction algorithm. Most of the obtained topologies are redundant and confusing, resulting in a complex graph and difficult operation in the subsequent search process. Moreover, existing model matching-based methods suffer from falling into the local optimum of the energy function.

To overcome these limitations, this paper presented a greedy graph search-based method for tracking vascular structures in angiographic image sequences guided by the vascular annotation in the key frame. In this work, each vascular branch is separated from the vasculature and tracked independently. Then, all branches are combined using topology optimization. This approach comprises two main steps. Greedy graph searching on an undirected acyclic graph was proposed, and branch matching via dynamic time warping was applied to produce the tracking results. Intensity-based image registration method was applied to two consecutive frames, and a deformation field was calculated to determine the tracking range. Then, the centerlines of the vasculature within the tracking range was extracted. In view of the high accuracy of the extraction of the vascular centerline in a single angiography image, the full use of the extracted vascular centerline effectively compensated for the drawback that the centerline tracked by previous methods was not at the center of the blood vessel. To enhance the accuracy of the graph construction, topology optimization was adopted to connect the gaps in the initial centerline results, thereby greatly reduce the time and storage space required for the branch search. An undirected acyclic graph was built based on these centerlines, and all possible paths in the graph that represented the branches of the vasculature were searched with candidate starting and stopping points. This kind of greedy thought can obtain the global optimal solution for vascular tracking. Therefore, this approach can effectively solve the problem that existing methods always fall into the local optimal solution. To exploit the image textural information corresponding to the vascular branch, a DAISY descriptor [29] was used to describe the characteristics of the tracked vascular branch in the key frame and all candidate vascular branches in the current frame. Dynamic time warping (DTW) was used to calculate the similarity between them, and the best matching vascular branch was retained and considered the branch tracking result. The complete vasculature tracking results were obtained by combining all branches through topology optimization.

## 2. Methods

In this study, a key frame with an initial vascular annotation was first selected from the specific angiographic sequence, and the image after the key frame was set as the current frame. All vascular branches of the initial annotation were split, and each branch was independently tracked on the current frame. The entire vasculature tracking result on the current frame was obtained by combining all of the vascular branch tracking results. Then, the current frame was set as a new key frame, and the subsequent frame image was set as the new current frame. A new round of tracking was proceeded until the vasculatures of all angiographic images in the sequence were labeled. In the beginning, the key frame is the first frame. In the Methods section, we used the $t$th frame as the key frame and $(t + 1)$th frame as the current frame to describe the algorithm. The key frame ($t$th frame) of a specific angiographic sequence was expressed as $A^t(x, y)$, and the current frame ($(t + 1)$th frame) was expressed as $A^{t+1}(x, y)$. The vascular annotation in the key frame was denoted as $V^t = \{(x_{11}, y_{11}), (x_{12}, y_{12}), \ldots, (x_{1M}, y_{1M}), (x_{21}, y_{21}), \ldots, (x_{2N}, y_{2N}), (x_{31}, y_{31}), \ldots, (x_{3P}, y_{3P})\}$, where $V^t$ indicates that three vessel branches are labeling the $t$th frame, and the numbers of points on the three branches were $M$, $N$, and $P$.

## *2.1 Pre-processing*

In this study, the vascular tracking problem was transformed into a graph-based vascular branch matching problem. Pre-processing is necessary before graph construction to restrain the tracking range in the space, thereby reducing the redundancy of the graph as well as the complexity of time and storage during the graph search.

*1) Image registration:* In many filamentary structural tracking methods, registration methods are typically used as the main pre-processing step. Regardless of type (i.e., image-based registration or skeleton-based registration), the purpose of this step is to map the vascular annotation of the key frame to the current frame to reduce the tracking range. We used Elastix [30] to implement an image-based registration of the key frame and the current frame. Thus, the blood vessel in the key frame was mapped to the current frame by using the registration deformation field.

*2) Vessel segmentation and centerline extraction:* Many methods in filamentary tracking from angiographic sequences, including the algorithm presented here, rely on binary segmentation masks and vessel centerlines. In an automatic processing pipeline, inspired by retinal segmentation, we applied a learning-based method [31, 32] to generate vascular segmentation results. To track each vascular branch, a neighborhood $\sigma$ was provided and used to lock the specific tracking range with the mapping position of the vascular annotation in the current frame. In this range, we applied multi-scale filtering and multi-directional NMS, which yield high responses across central positions and smooth vessel centerlines, respectively. Details on the automatic segmentation and centerline extraction algorithms used here can be found in [21, 33]. However, the methods discussed in the remainder of this paper are not specific to particular segmentation and centerline extraction algorithms.

## *2.2 Graph Building and Greedy Branch Searching*

Given that the general centerline extraction algorithm yields discontinuous results, we applied a multi-probability fusion centerline topology optimization method to repair the gaps. A connection probability map was constructed by the tensor field formed by the initial centerline skeleton as well as the textural and orientation information of the vasculature. The discontinuous range was repaired by using the Dijkstra algorithm to search for the optimal connection path among the gaps. Consequently, complete centerline results were obtained (as illustrated in Fig 2(a) and Fig 2(b)).

Centerline segments were obtained by splitting the complete centerlines by using the bifurcation points **B**. Each segment was a polygonal curve $C(i)$ on the plane. The graph $\boldsymbol{G} = (\boldsymbol{C}, \boldsymbol{B})$ was defined to describe the relationship between centerline segments $\boldsymbol{C}$ and bifurcation points $\boldsymbol{B}$, where the edges corresponded to the centerline segments, and the nodes corresponded to the bifurcation points. The graph contains two kinds of connection relations: 1) the connection between two ends belonging to the same centerline curve; and 2) the connection between two ends belonging to different centerline curves but sharing the same bifurcation point. Based on the connection relationship of the nodes in the graph, an adjacency matrix $G$ can be defined as follows:

$$G(k,l) = \begin{cases} 1 & \text{if } n_k \text{ and } n_l \text{ are connected} \\ 0 & \text{if } n_k \text{ and } n_l \text{ are not connected} \end{cases} \quad (1)$$

An undirected acyclic graph was established, as shown in Figs.2(c)–2(f). A complete vascular branch can be seen as a path from a starting point to an ending point on the graph. Therefore, as long as the starting and ending points that might match the endpoints of the vascular branch in the key frame are given, we can search all paths that might match with the guided vascular branch.

In two successive frames of an angiographic sequence, the endpoint displacement of the mapped guided branch and the real vascular branch was not significant. As such, we used a spatial distance constraint to design an endpoint selection strategy. In the current frame, the candidates {*can_start*} of the starting point of the tracked branch were selected by retaining the endpoints of the segments, which consisted of the nearest *n* segments from the guided starting point. Similarly, the candidates {*can_end*} of the ending point were selected (as illustrated in Fig 2(g)). All possible tracking results were non-looped paths in graph $\boldsymbol{G}$ that were identified using the depth-first search algorithm [34]. Their starting and ending points were in {can_start} and {can_end}, respectively.

We defined the possible tracking results as $V'^{t+1}_m = \{V'^{t+1}_{1m}, V'^{t+1}_{2m}, ..., V'^{t+1}_{Qm}\}$, which denotes that Q possible results are available for the $m$th branch of the vasculature on the $(t+1)$th frame of the specific angiographic sequence, where $V'^{t+1}_{Qm} = \{(x_1, y_1), (x_2, y_2), ..., (x_L, y_L)\}$ represents that the $Q$th possible result contains $L$ points. The complete process of graph building and greedy branch searching is shown in Fig. 2.

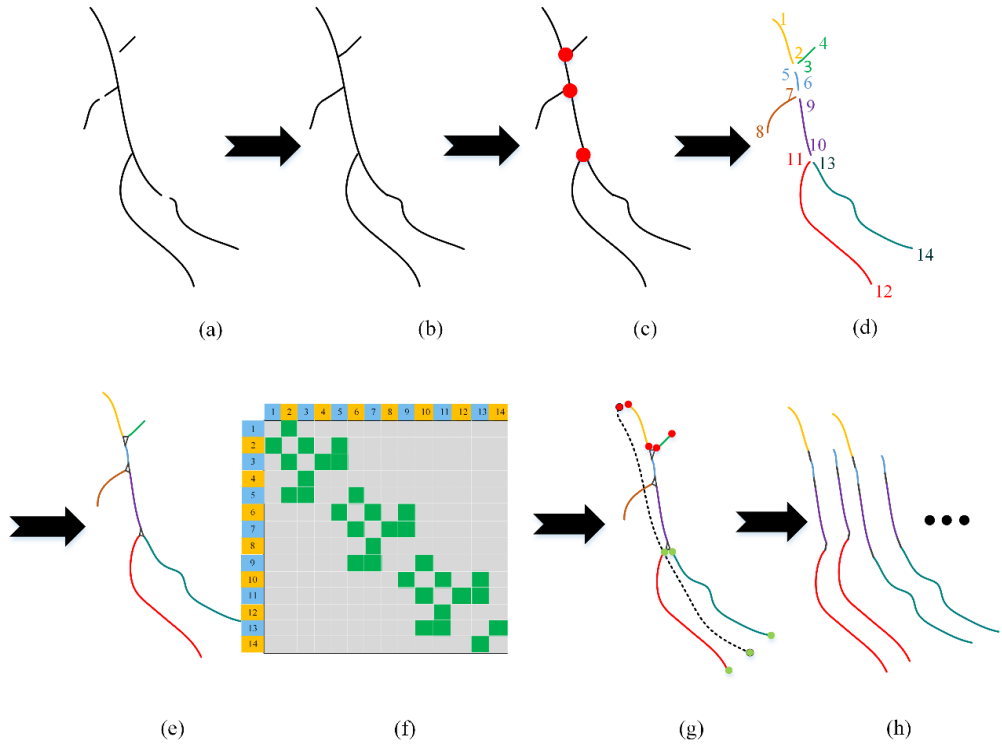

Fig.2 The process of graph building and greedy branch searching. (a) initial centerlines; (b) repaired centerlines; (c) bifurcations (red points); (d) vessel segments divided using bifurcations; (e) judgment of connection relations; (f) adjacency matrix G; (g) selection of the starting and ending points; (h) possible tracking results of a specific branch.

## 2.3 Branch Matching via Dynamic Time Warping

Among all possible tracking results, the final result was the branch that was most similar to the guided branch. Assuming that the guided branch was represented as $V^t_m$, with regard to the length of the vascular branch, $V'^{t+1}_{Q_m}$ is not always equal to $V^t_m$ because of cardiac contraction and expansion. In this study, dynamic time warping (DTW) algorithm was used to calculate the match between two branches with different lengths. DTW is a pattern matching algorithm based on nonlinear dynamic programming that calculates the relationship between various time points [28, 35]. As shown in Fig.3 (a), two sequences on the left and right represented branch $V^t_m$ whose length was $M$ pixels and branch $V'^{t+1}_{Q_m}$ whose length was $L$ pixels. The dashed lines between the two sequences represented the matched point pairs. DTW exploits the warping distance to measure the difference between two sequences. This distance was obtained by calculating the difference between every matched point pair. A smaller warping distance corresponded to a greater similarity of the two sequences. Warping distance depends on the warping path. A warping path can be expressed as $W = \{w_1, w_2, \ldots, w_k, \ldots, w_K\}$, where $max(M, L) \leq K \leq M + L$, and $w_k = (i, j)$, where $i$ is the $i$th point in $V^t_m$, and $j$ is the $j$th point in $V'^{t+1}_{Q_m}$. $w_k$ denotes that two points are a matched point pair. Every $w_k$ corresponded to a distance $d_k(i, j)$, which was the difference between a specific point pair. In this study, the warping path $W$ must satisfy three conditions.

$$\begin{cases} (1) \ w_1 = (1,1), w_K = (M, L) \\ (2) \ w_k = (i, j), w_{k+1} = (i', j') \quad i' - i \leq 1, j' - j \leq 1 \\ (3) \ w_k = (i, j), w_{k+1} = (i', j') \quad i' - i \geq 0, j' - j \geq 0 \end{cases} \quad (2)$$

The first constraint is boundary condition, which guarantees that the alignment points in two sequences start from the start point and end at the end point. The second condition is continuity, which guarantees that a point on a sequence can be aligned only with its neighboring point in another sequence, such that each point can appear in the warping path. The third constraint is monotonicity, which ensures that the dashed lines do not intersect (Fig. 3(a)). The objective function of the warping path can be defined as follows:

$$DTW_{path}\left(V_m^t, V'^{t+1}_{Q_m}\right) = W^* = \min_{W}\{\sum_{k=1}^{K} d_k\} \quad (3)$$

$$DTW_{distance}\left(V_m^t, V'^{t+1}_{Q_m}\right) = \min\{\sum_{k=1}^{K} d_k\} \quad (4)$$

This problem can be solved by dynamic programming (DP) [36]. As shown in Fig. 3(b), a graph $O$ was created, where each grid point represented the corresponding point pair of two vascular branches $[V_m^t(i), V'^{t+1}_{Q_m}(j)]$, and is denoted as $O(i,j)$. Finding all of the corresponding point pairs in two branches involves finding a path from $O(0,0)$ to $O(M-1, L-1)$ and connecting all of the matched point pairs in graph G. The cost value of node $O(i,j)$ is $d(i,j)$. The warping distance can be regarded as a cost value. Thus, we constructed a cost matrix $D$ with $M \times L$ size. $D(i,j)$ represented the cumulative cost of the optimal warping path under the current length, and the current length is the vascular segment with $i$ points $V_m^t(0:i)$ and the vascular segment with $j$ points $V'^{t+1}_{Q_m}(0:j)$. $D(i,j)$ can be calculated as follows:

$$D(i,j) = d(i,j) + \begin{cases} 0 & if\ i=0, j=0 \\ D(i, j-1) & if\ i=0, j>0 \\ D(i-1, j) & if\ i>0, j=0 \\ \min(D(i-1,j), D(i,j-1), D(i-1,j-1)) & if\ i>0, j>0 \end{cases} \quad (5)$$

Therefore, $D(M,L)$ is the warping distance between the holistic $V_m^t$ and $V'^{t+1}_{Q_m}$ under the optimal warping path. After the warping distance was calculated between $V_m^t$ and every branch in $\{V'^{t+1}_{1_m}, V'^{t+1}_{2_m}, \dots, V'^{t+1}_{Q_m}\}$, the final tracking result $V_m^{t+1}$ was selected by retaining the branch with the smallest warping distance.

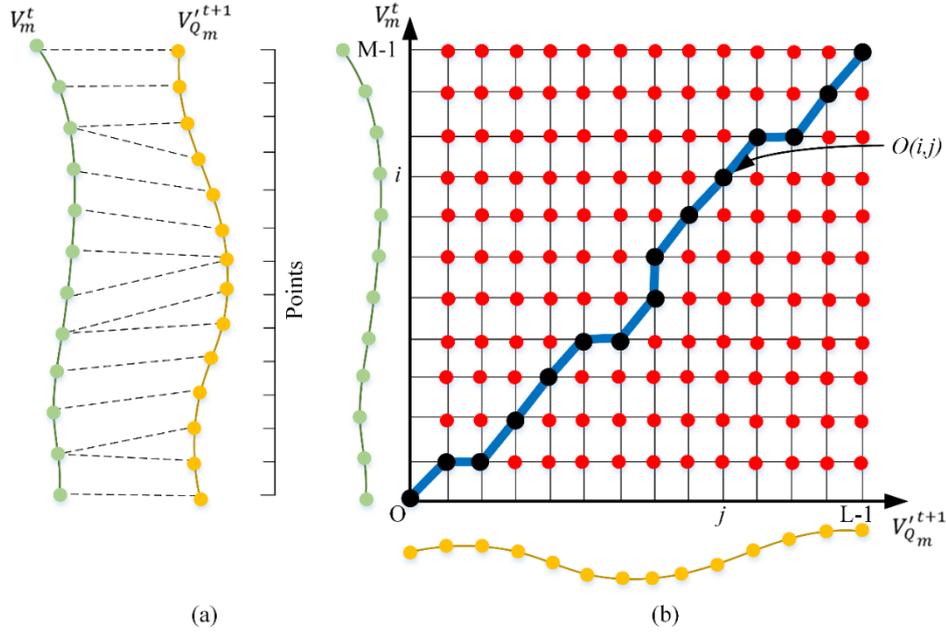

Fig.3 Schematic of DTW. (a) Corresponding point pairs in DTW; (b) searching for an optimal warping path with DP.

The cost value $d(i,j)$ corresponding to the node $O(i,j)$ reflected the difference in the vascular characteristics between the matched points, which is calculated as follows:

$$d(i,j) = F(i) \cdot F(j) \qquad (6)$$

where $F(\cdot)$ represents the image feature corresponding to the point on the vasculature. In this study, local feature descriptors were used to describe the image feature $F(\cdot)$, and the difference between two points can be easily calculated by measuring the distance between two feature description vectors. A DAISY descriptor [29] was employed for the feature to more effectively describe the points on the angiograms.

## 2.4 Datasets and Evaluation Metrics

Our method was evaluated on two datasets: single branch dataset (SBD) and a vessel tree dataset (VTD). SBD was collected to evaluate the performance of vascular branch tracking, whereas VTD was used to evaluate the performance of complex vasculature tracking. SBD contains 12 angiographic image sequences with 77 angiograms of contrast agent-filled vessels, and VTD contains 9 angiographic image sequences with 58 angiograms of contrast agent-filled vessels. Both datasets were provided by Anzhen Hospital, Beijing. Here, the image resolution was 512 × 512 pixels and the frame rate for all of these angiographic sequences was 15 frames/s. Three experts manually delineated the centerlines in the 135 clinical angiograms, and the averaged centerline results were used as the ground truth.

In this study, precision (*Prec*), sensitivity (*Sens*), and F1 score (*F1*) were used to measure the performance of the proposed method. Precision refers to the accuracy of the tracking results. Sensitivity represents the completeness of the tracking results. The F1 score reflects both accuracy and completeness. Specifically, the chosen metrics were defined as follows: $Prec = TP/(TP + FP)$; $Sens = TP/(TP + FN)$ and $F1 = 2 \times Prec \times Sens/(Prec + Sens)$, where *TP*, *FP*, and *FN* represent true positives, false positives, and false negatives, respectively. In the quantitative analysis of the performance of the algorithm, the mean and standard deviation (STD) of the selected metrics were calculated, as the mean reflected the performance trend, whereas the STD denoted the stability. Similar to the evaluation methods for extracting one-pixel-wide curves, we introduced a tolerance factor $\rho$ [14] to calculate the three metrics.

## 3. Experimental Results

### *3.1 Results on Different Datasets*

This section examines the performance of our algorithm in tracking a single vascular branch on SBD and a holistic vasculature on VTD. In Fig.4, the first two rows are the tracking results of four examples in SBD, and the last two rows are the results in VTD. In the figure, (a) and (d) show the guided frame image of a sequence, and the yellow centerlines represent the vascular annotation to be traced; (b) and (e) show an angiographic image in the sequence; and the green centerlines in (c) and (f) are the tracking results of the guided branch or vasculature. The tracking results of our algorithm can completely and accurately cover the vessels to be tracked in both datasets. Table 1 provides the quantitative evaluation results in terms of precision, sensitivity, and F1 score. Our method achieved $Prec = 0.90 \pm 0.07$, $Sens = 0.89 \pm 0.08$, and $F1 = 0.89 \pm 0.06$ for SBD and $Prec = 0.89 \pm 0.05$, $Sens = 0.87 \pm 0.05$, and $F1 = 0.88 \pm 0.05$ for VTD. The high mean and low STD values confirmed the effectiveness and stability of our algorithm.

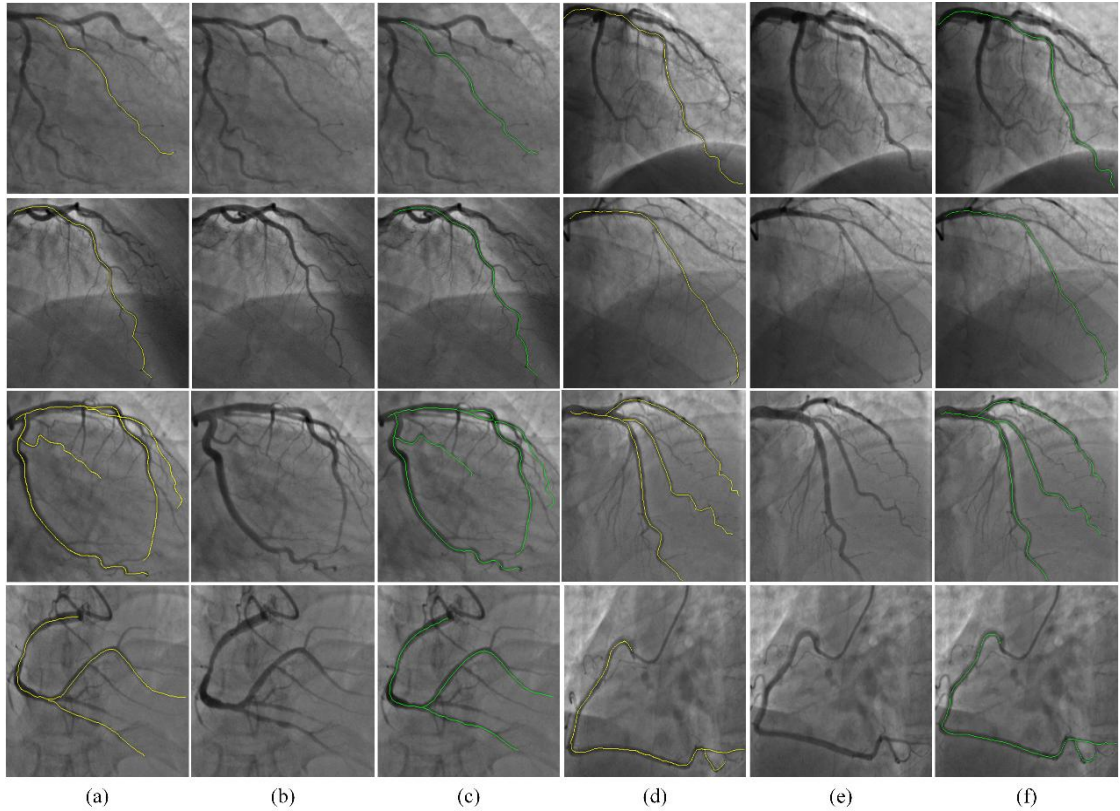

(a)　　　　(b)　　　　(c)　　　　(d)　　　　(e)　　　　(f)

Fig.4 The vascular tracking results of the proposed method on the instanced angiographic images of the sequence in two types of datasets. First two rows: results on SBD, last two rows: results on VTD. (a) (d): first frame angiogram of a sequence and the vascular branch or vessel tree to be tracked (yellow centerlines); (b) (e): a certain frame

angiogram of the sequence; (c) (f): the corresponding vascular tracking results (green centerlines).

Table I. The performances (mean±STD) of the proposed method on vascular tracking over different datasets at the pixel level

| Dataset | *Prec* | *Sens* | *F1* |
|---------|--------|--------|------|
| **SBD** | 0.90±0.07 | 0.89±0.08 | 0.89±0.06 |
| **VTD** | 0.89±0.05 | 0.87±0.05 | 0.88±0.05 |

The tracking accuracy is affected by the tracking span because the tracking task on the next angiogram is guided by the tracking result of the previous frame. Fig. 5 shows the vascular tracking results of the first, middle, and last frames in four sequences on two datasets. The first two rows are the results on SBD, and the last two rows are the results on VTD. In the figure, (a) shows the first frame angiograms of a certain sequence and its vascular annotation to be tracked (yellow centerlines); (a1)–(a3) are the first, middle, and last frame angiograms in the sequence; the green centerlines in (b1)–(b3) are the corresponding tracking results. As shown, the vascular tracking results of our algorithm were accurate regardless of the frame of the sequence, indicating that the number of tracked frames insignificantly affected our proposed method. The F1 score of the tracking performance in the first, middle, and last frames were 0.91, 0.91, and 0.86 on SBD and 0.94, 0.88, and 0.82 on VTD. In line with the conclusion drawn from the qualitative results, the quantitative results verified that the tracking accuracy of our algorithm was insignificantly affected by the tracking span. In addition, the proposed method can well handle angiograms with considerable background noise (row 1 in Fig. 5), substantial changes in the vascular curvature (row 2 in Fig.5), low contrast between the vasculature and the background (row 3 in Fig. 5), and significant changes in the vascular radius (row 4 in Fig. 5).

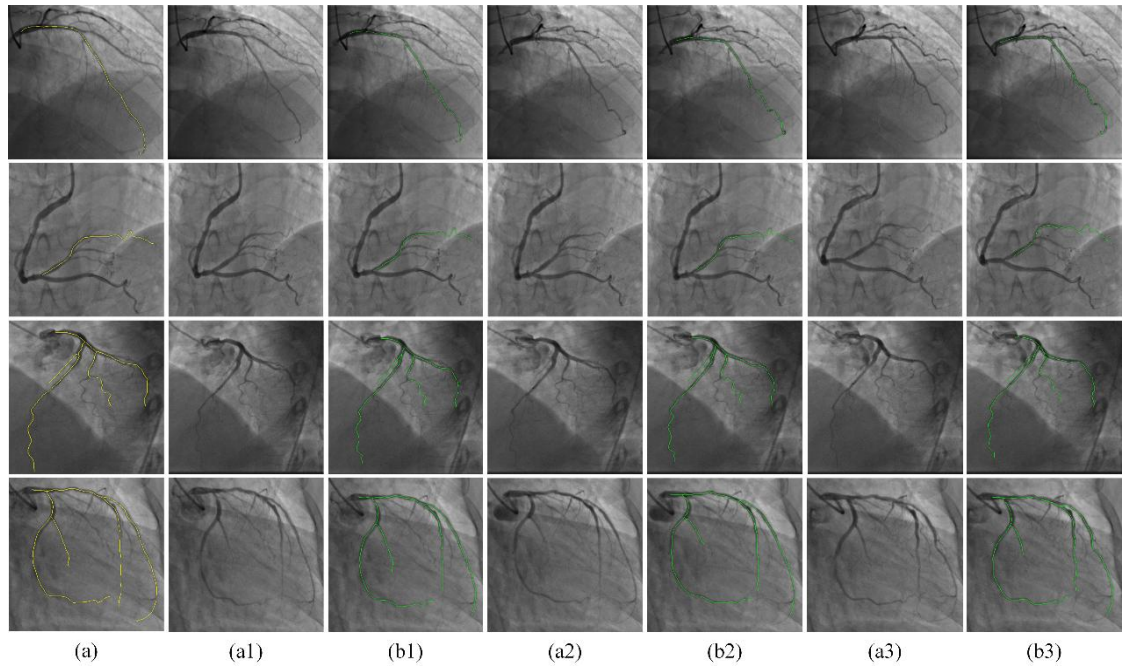

    (a)        (a1)        (b1)        (a2)        (b2)        (a3)        (b3)

Fig.5 vascular tracking results in four certain sequences on two datasets. First two rows: results on SBD; last two rows: results on VTD. (a): first frame of a sequence and its vascular annotation to be tracked (yellow centerlines); (a1)-(a3): the first, middle and last frames angiograms in the sequence; (b1)-(b3): corresponding vascular tracking results (green centerlines).

## 3.2 Results on Parameter Setting

Our proposed method included the following parameters: the size $\sigma$ for determining the tracking range in the current frame, the number of nearest neighbor segments $n$ for selecting the starting and ending points, and whether topology optimization is sued to combine branches. In this section, we experimentally investigated their effects of these parameters on the tracking results.

Parameter $\sigma$ was applied in the pre-process as a spatial constraint. This parameter is affected by the amplitude of the vascular motion in two successive frames: a larger motion amplitude corresponded to a larger parameter. All angiographic image sequences in our datasets were acquired from people with a normal heart rate (75 beats/min), and every 12 frames constituted a heartbeat period. We chose $\sigma = 5$ pixels in our experiments. In view of the assumption that the sequences were acquired from people with heart rate of 120 beats/min, a heartbeat period constituted of only approximately 7 frames. The first and third frames of each sequence in the existing datasets simulated the first two consecutive frames images in the sequences of the patient with tachycardia. To test the tracking performance of our algorithm on the sequences with large vascular motion, we designed an experiment that can track the vasculature in the third frame image by using the vascular annotation in the first frame as guide. We tested 21 sequences on the two datasets, and we chose $\sigma = 25$ pixels. The quantitative evaluation of the results were *Prec* = 0.93 ± 0.03, *Sens* = 0.91 ± 0.08, *F1* = 0.92 ± 0.06 on SBD and *Prec* = 0.89 ± 0.09, *Sens* = 0.88 ± 0.06, *F1* = 0.88 ± 0.07 on VTD. If the vascular motion of two consecutive frames is large, the validity and stability of the algorithm can be ensured by increasing $\sigma$.

In searching for the possible branches in the graph, the possible starting and ending points were selected by retaining the endpoints of the segments, which were the nearest *n* segments from the guided endpoints. A suitable *n* value must be selected, as a small value would lead to an incomplete search of possible endpoints, whereas a large value would increase the search time and the storage space. In this section, we experimentally tuned this parameter by adjusting it to 1, 2, and 3. Fig. 6 shows the performances of the proposed method under these conditions and reveals that the proposed method achieved the best performance when $n = 2$. Here, (a) and (b) present the evaluation results of three metrics on SBD and VTD, respectively.

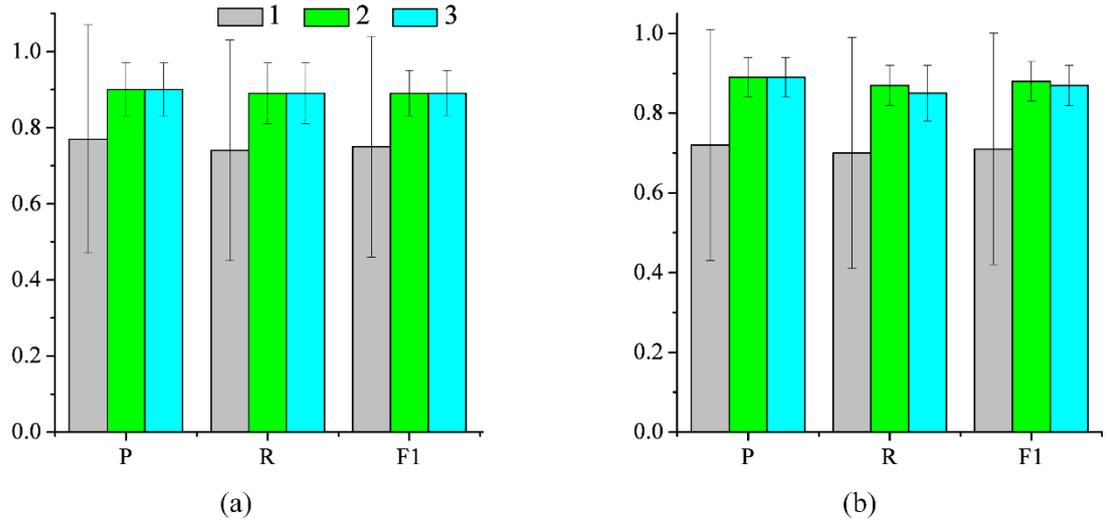

Fig.6 Influence of searching range of possible starting and ending points on algorithm performance. (a) result on SBD; (b) result on VTD

To obtain the vascular tree tracking results with a continuous and complete topology information, we used a topology connection optimization algorithm to combine each vascular branch. The following experiment verified the effect of using the optimization algorithm. Table II provides the quantitative results. As shown, incorporating the topology connection optimization operation increased the tracking result to a certain extent: *Prec* increased by 2%, and *F1* increased by 1%.

Table II. Mean ± STD of various metrics of the proposed method with and without using topology optimization

| Topology Optimization | *Prec* | *Sens* | *F1* |
|---|---|---|---|
| **Without** | 0.89±0.05 | 0.85±0.06 | 0.87±0.05 |
| **With** | 0.89±0.05 | 0.87±0.05 | 0.88±0.05 |

### *3.3 Results on Comparing*

In this section, we conducted a comparative study between the proposed method and the state-of-the-art ones for vascular tracking on SBD and VTD. To this end, five filamentary structure tracking methods were selected: image registration-based method [30], deformable model-based method [16], point set registration-based method [37], general object tracking-based method [38], and vascular key point tracking-based method [21]. The competitors and the proposed method were referred to as Elastix, OpenSnake, GMM, SPOT, VCO and Proposed. Corresponding to the classes of tracking methods described in Section I, Elastix is an image registration method; GMM and Proposed are similar to prior model matching methods; OpenSnake, SPOT, and VCO are motion tracking methods. The source codes were used with the default parameter settings provided by the authors who proposed these methods. The tracking results obtained by GMM on our datasets had the problem of interruption during tracking in the sequence, suggesting that this method failed to perform vasculature tracking in angiographic sequences. The performance analysis of the other four compared methods and the proposed method is provided as follows.

The vascular tracking results on two datasets obtained by five different methods are presented in Fig.7. Here, (a) shows the guided frame image of a sequence and the vascular annotation to be traced (yellow centerlines); (b) shows an angiographic image in the sequence; (c)–(g) show the vascular tracking results of the appointed angiogram obtained by Elastix, OpenSnake, SPOT, VCO, and Proposed, respectively. Red indicated the regions with significant errors, and green indicates those with minimal errors. As shown, the proposed method can successfully obtain accurate vasculatures on either SBD or VTD. Elastix used the image information to obtain a deformation field of two consecutive frames and then mapped the guided vascular annotation. Although the obtained result can cover the vascular structure (Fig. 7(c)), the tracked centerline was unsmooth (Fig.

8(c)). OpenSnake provided a deformation model for the vascular structure with the use of the image intensity information. Tracking results with significant errors were often obtained because of the difficulty of setting the energy function and iteration parameter (Fig. 7(d)). SPOT used a target tracking algorithm to track several key points on the blood vessel and a livevessel algorithm [39] to connect the tracked key points to ultimately obtain the vascular tracking results. This algorithm obtained poor results because the tracking of the key points was not robust (Fig. 7(e)). VCO is similar to SPOT but was designed specifically for vascular tracking in angiographic image sequences. VCO used Markov to optimize the target tracking results and FastMarching to obtain the connection among target points. The vascular tracking results occasionally have wrong branches because of the deviation of the target tracking results (Fig. 7(f)). Compared with other methods, the proposed method can not only track accurate and complete vasculatures but also attain smooth centerlines.

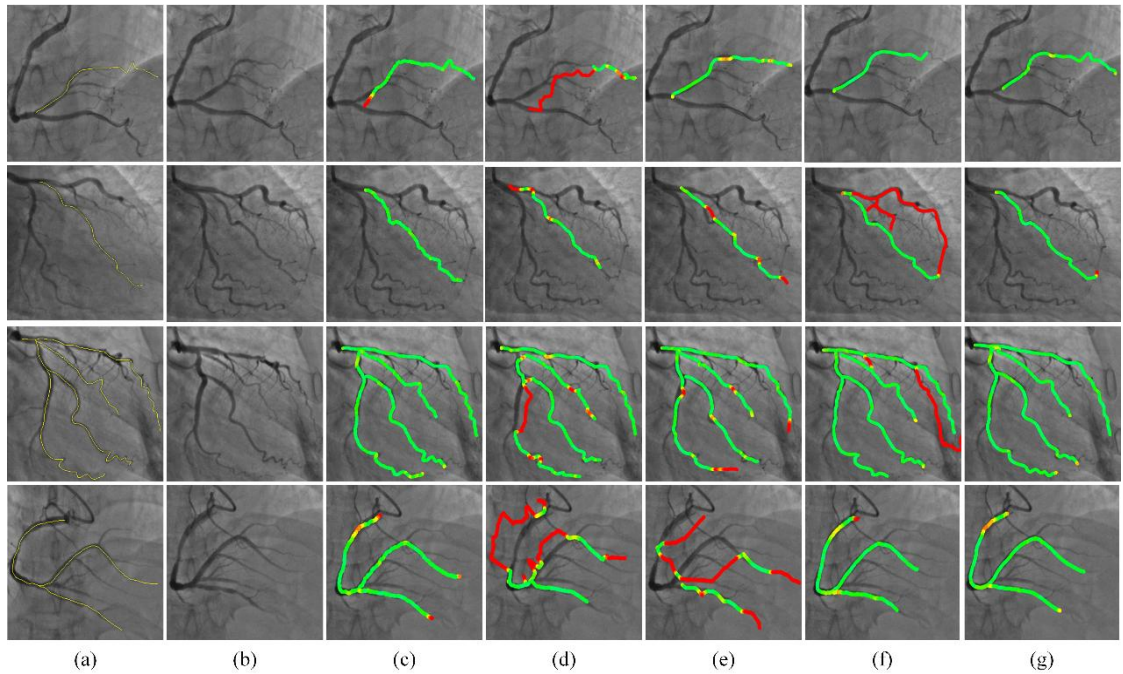

(a) (b) (c) (d) (e) (f) (g)

Fig. 7 Vascular tracking results on two datasets obtained by five different methods. First two rows: results on SBD;

last two rows: results on VTD. (a): first frame of a sequence and its vascular annotation to be tracked (yellow centerlines); (b): a certain angiogram in the sequence; (c)-(g): vascular tracking results obtained by Elastix, OpenSnake, SPOT, VCO and Proposed, respectively. Red indicates regions with large errors, while green indicates small errors.

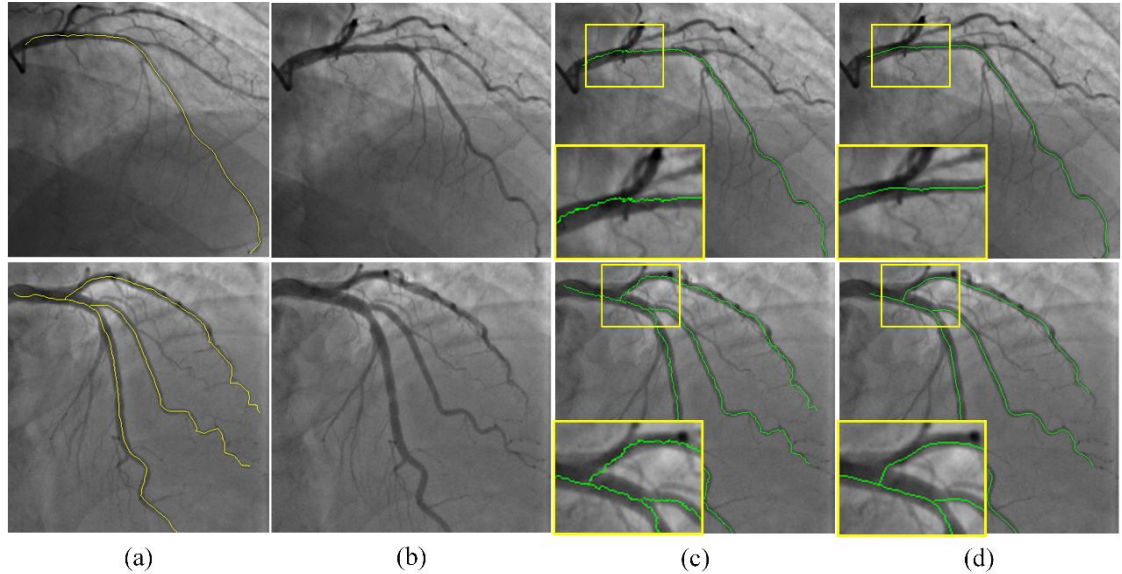

(a)　　　　　　　(b)　　　　　　　(c)　　　　　　　(d)

Fig.8 Comparison of the vascular centerlines obtained by Elastix and Proposed. (a): the guided angiogram and its vascular annotation (yellow centerlines); (b): a certain angiogram in the sequence; (c): centerline results tracked by Elastix; (d): centerline results tracked by Proposed. Rectangle in yellow is the zoom-in view of a selected region.

To objectively measure the performance, the precision, sensitivity, and F1 score of the vascular tracking results derived by different methods were calculated. Table III presents the average results obtained by using our method and its competitors over 77 angiograms on SBD and over 58 angiograms on VTD. Our method achieved the best performance with $Prec = 0.90 \pm 0.07$, $Sens = 0.89 \pm 0.08$, and $F1 = 0.89 \pm 0.06$ over SBD. Among the other methods, VCO displayed the best performance with $Prec = 0.87 \pm 0.13$, $Sens = 0.80 \pm 0.18$, and $F1 = 0.81 \pm 0.14$. The higher mean and lower STD values verified that our algorithm was superior to the other four algorithms in tracking a single vascular branch. Over VTD, the quantitative results suggested that Elastix was the best. However, as discussed, the vascular centerline tracking results obtained by this algorithm are only stacks of individual pixels and therefore are unsmooth and have no connection information among points. Although the performance of our method was inferior to Elastix on VTD, the tracking results were consistent with the characteristics of the vascular centerlines, making it more suitable for application in vascular tracking tasks. In addition, the proposed method was significantly better than the other three methods in the quantitative results. Therefore, our algorithm outperformed the other four state-of-the-art methods for vascular tracking on both SBD and VTD.

Table III. The performances (mean ± STD) of different methods over SBD and VTD

| Dataset | SBD | | | VTD | | |
|---|---|---|---|---|---|---|
| Method | *Prec* | *Sens* | *F1* | *Prec* | *Sens* | *F1* |
| **Elastix** | 0.76±0.26 | 0.54±0.35 | 0.59±0.32 | 0.88±0.05 | 0.89±0.04 | 0.88±0.04 |
| **OpenSnake** | 0.68±0.16 | 0.49±0.28 | 0.52±0.32 | 0.72±0.13 | 0.74±0.04 | 0.73±0.10 |
| **SPOT** | 0.67±0.22 | 0.67±0.19 | 0.66±0.21 | 0.70±0.17 | 0.67±0.16 | 0.68±0.16 |
| **VCO** | 0.87±0.13 | 0.80±0.18 | 0.81±0.14 | 0.86±0.07 | 0.82±0.06 | 0.84±0.05 |
| **Proposed** | 0.90±0.07 | 0.89±0.08 | 0.89±0.06 | 0.89±0.05 | 0.87±0.05 | 0.88±0.05 |

Similar to Fig. 5, Figs. 9 and 10 qualitatively show the relationship between the tracking accuracy and tracking span of different methods in SBD and VTD, respectively. In the figure, the rows from top to bottom correspond to the tracking results of the first, middle, and last frames of a certain sequence, respectively. (a) shows a certain angiogram, and (b)–(f) show the corresponding tracking results of Elastix, OpenSnake, SPOT, VCO, and Proposed, respectively. In view of the space limitations, the guided vascular structure is not shown here. As shown, the tracking results of Proposed and VCO were insignificantly affected, whereas those of OpenSnake and SPOT were significantly affected, by the tracking span. Fig. 11 shows the quantitative relationship between the tracking accuracy and the tracking span, where the horizontal axis represents the first, middle, and last frames of all sequences in the datasets, and the vertical axis represents the corresponding average tracking accuracy. Here, (a) and (b) show the results of SBD and VTD, respectively. The tracking accuracy of Elastix, VCO, and Proposed were insignificantly affected by the tracking span in both datasets, which was consistent with the qualitative results. Furthermore, the tracking accuracy of Proposed was the highest over SBD. Over VTD, given that the results tracked by Elastix failed to satisfy the characteristics of the vascular centerlines, the proposed method exhibited the best performance.

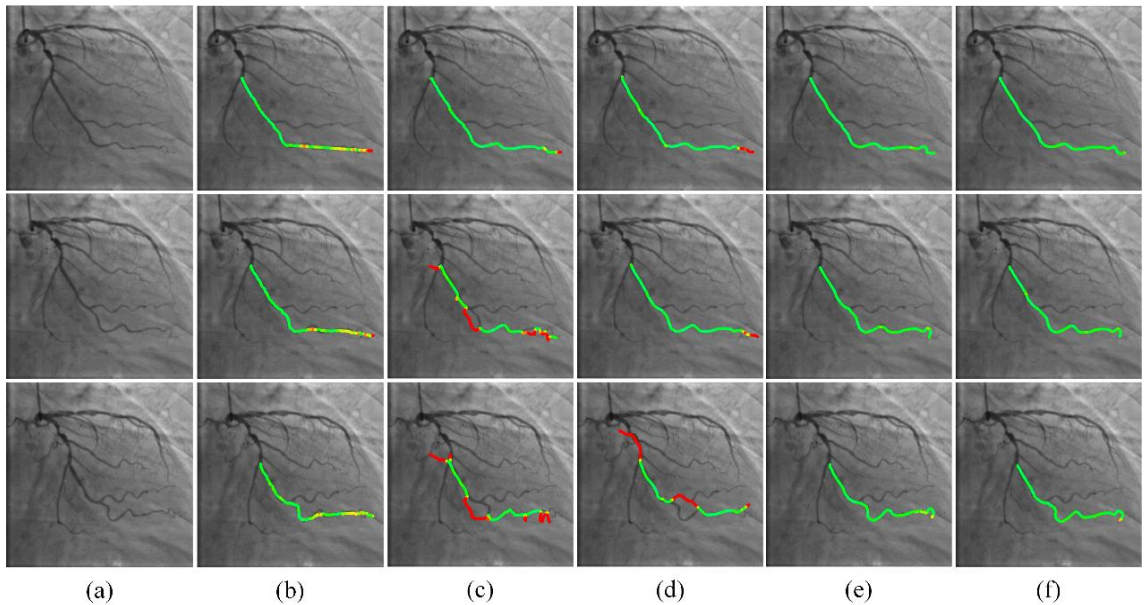

(a) (b) (c) (d) (e) (f)

Fig.9 Vascular tracking results of different methods in a certain sequence over SBD. Rows from top to bottom: tracking results on the first, middle and last frame of this sequence; (a): a certain angiogram of the sequence; (b)-(f): vascular tracking results obtained by Elastix, OpenSnake, SPOT, VCO and Proposed, respectively. Red indicates

regions with large errors, while green indicates small errors.

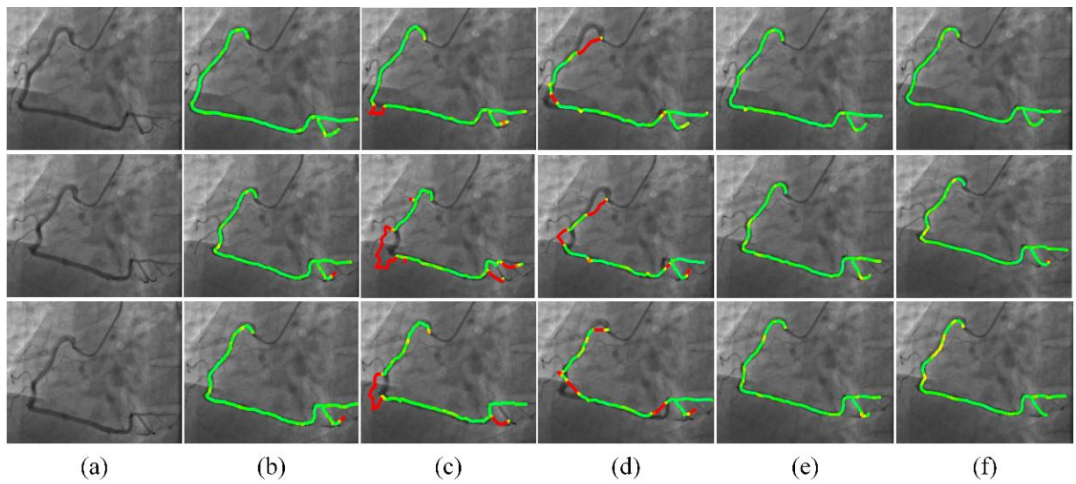

Fig.10 Vascular tracking results of different methods in a certain sequence over VTD. Rows from top to bottom: tracking results on the first, middle and last frame of this sequence; (a): a certain angiogram of the sequence; (b)-(f): vascular tracking results obtained by Elastix, OpenSnake, SPOT, VCO and Proposed, respectively. Red indicates regions with large errors, while green indicates small errors.

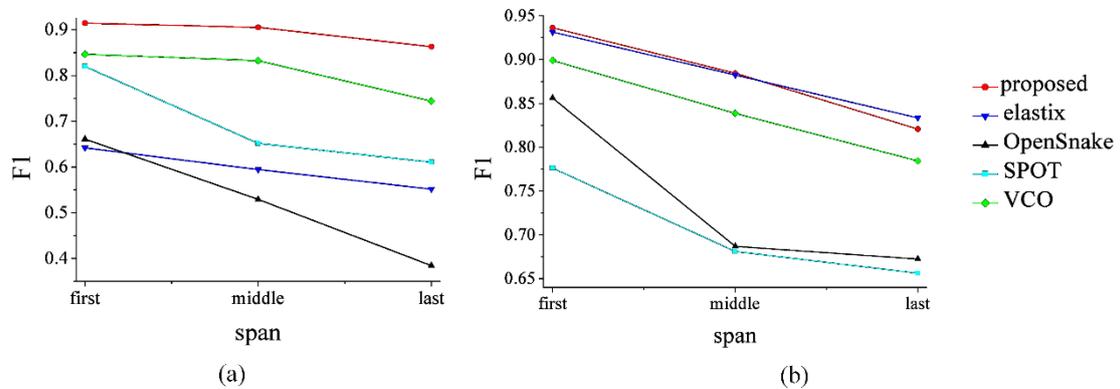

Fig.11 The relationship between tracking accuracy and tracking span. (a): result on SBD; (b): result on VTD

## 4. Conclusions

Extracting continuous coronary structures through X-ray angiographic image sequences is an important aspect in the computer-assisted treatment of vascular diseases. Such an approach can provide vascular structural information for coronary motion estimation, 4D reconstruction of coronary arteries, X-ray-guided cardiovascular interventional surgery, and so on, all of which can assist in the analysis of vascular status, degree of disease, intraoperative display, and other issues. In this paper, we translated the problem of the vasculature extraction of sequence images into a vasculature tracking problem based on the known vascular structures in the first frame. We exploited both the spatial and textural information between successive frames. Then, we introduced a novel vasculature tracking method based on greedy graph searching and branch matching.

In this method, the extracted vascular centerlines from every single angiogram were fully used to avoid the drawback that the centerline tracked by previous methods was not at the center of the blood vessel. In addition, topology optimization was applied to repair the gaps in the extracted centerlines and in turn enhance the accuracy of the graph construction. In this way, the time and storage space required for branch search can be greatly reduced. Greedy graph search was proposed to generate the global optimal solution for the tracking problem. This technique can effectively avoid the problem that existing methods occasionally fall into a local optimal solution. Branch matching using DTW with a DAISY descriptor was applied to determine the final tracking result, because the DAISY descriptor can describe the characteristics of vascular branches by utilizing the image textural information, and DTW can calculate the matching of two branches with different lengths.

The proposed method was validated using clinical datasets, and the results are robust, demonstrating the effectiveness of our proposed approach. The tracked centerline results well matched the characteristics of the blood vessels. Qualitative and quantitative analysis results revealed that the proposed approach clearly outperformed the other state-of-the-art algorithms.

Many other clinical relevant applications can potentially benefit from our result. The results could help establish the relationship between the structure of vasculature and pathologies to better understand its physiology. The underlying technologies from our methods are not limited to vasculature tracking and can easily be generalized to track other tubular structures (e.g., guide-wire). In future works, we plan to use the vascular topological information to track all vascular branches in the vasculature simultaneously.

## Conflict of Interests

The authors declare that they have no competing interests.

## Acknowledgment

This work was supported by the National Key Research and Development Program of China (2017YFC0107900), and the National Science Foundation Program of China (61672099, 81627803, 61501030, 61527827).